\documentclass{article}
\usepackage{authblk}
\usepackage[utf8]{inputenc}
\usepackage{url}
\usepackage{hyperref}
\usepackage{color}
\usepackage{nips12submit_e,times}
\usepackage{booktabs}
\usepackage{graphicx}
\usepackage{caption}
\usepackage{subcaption}

\title{Fine-grained Categorization -- Short Summary of our Entry for the ImageNet Challenge 2012}
\author[1]{Christoph Göring}
\author[1]{Alexander Freytag}
\author[2]{Erik Rodner}
\author[1]{Joachim Denzler}
\affil[1]{Computer Vision Group, FSU, Jena, Germany}
\affil[2]{ICSI Vision Group, UC Berkeley, California}

\begin{document}

  \maketitle

\begin{abstract}
In this paper, we tackle the problem of visual categorization of dog breeds, which is 
a surprisingly challenging task due to simultaneously present  low interclass distances and high intra-class variances.
Our approach combines several techniques well known in our community but often not utilized for fine-grained recognition:
 (1) automatic segmentation, (2) efficient part detection, and (3) combination of multiple features. 
In particular, we demonstrate that a simple head detector embedded in an off-the-shelf recognition pipeline
 can improve recognition accuracy quite significantly, highlighting the importance of part features for fine-grained recognition tasks.
Using our approach, we achieved a $24.59 \%$ mean average precision performance on the Stanford dog dataset. 
\end{abstract}

\section{Introduction}
\begin{figure}[b]
        \centering
        \begin{subfigure}[b]{0.49\textwidth}
                \centering
\includegraphics[height=0.27\textwidth]{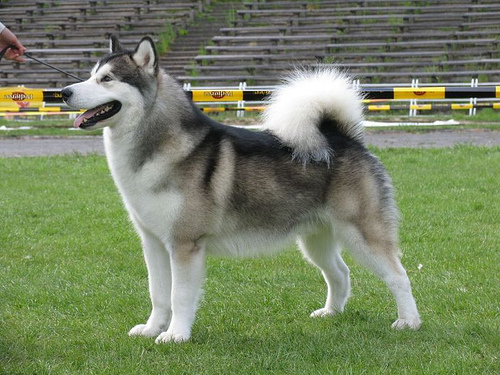}
\includegraphics[height=0.27\textwidth]{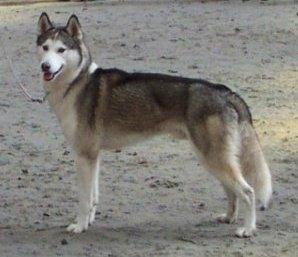}
\includegraphics[height=0.27\textwidth]{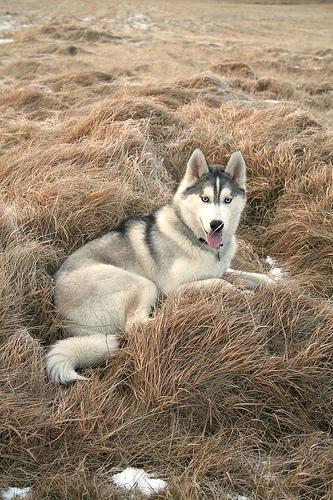}
                \caption{different breeds of huskies}
				\label{fig:huskies}
        \end{subfigure}
        \begin{subfigure}[b]{0.49\textwidth}
                \centering
\includegraphics[height=0.27\textwidth]{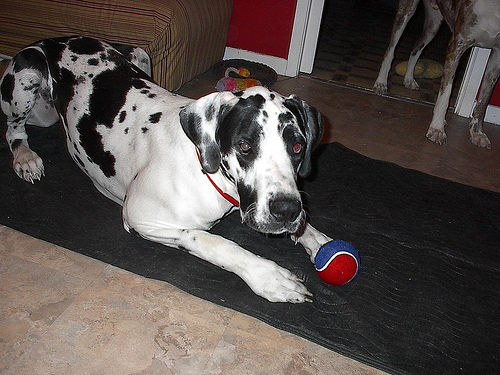}
\includegraphics[height=0.27\textwidth]{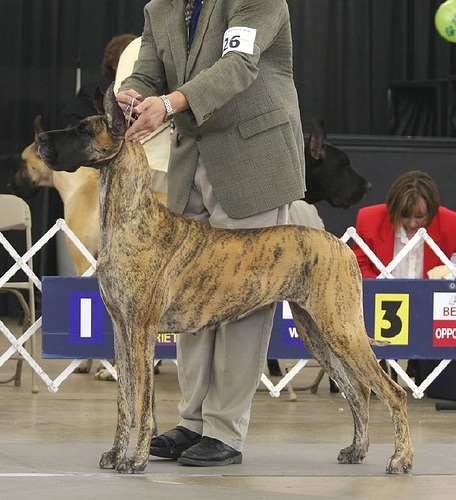}
\includegraphics[height=0.27\textwidth]{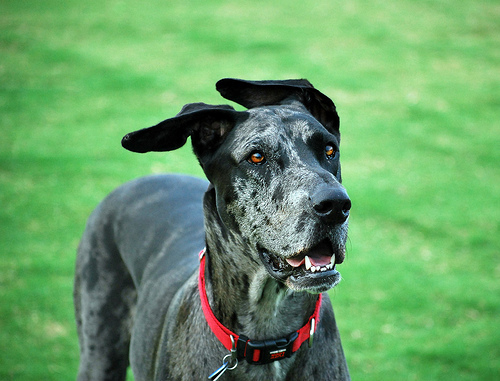}
                \caption{Great Dane fur color variation}
				\label{fig:greatdanes}
        \end{subfigure}
\caption{The problems associated with fine-grained classification: (\subref{fig:huskies}) The interclass variance can be low whereas the (\subref{fig:greatdanes}) intra-class variance can be very high. Together with the relatively large number of classes this leads to a very challenging task.}
	\label{fig:finegrainedproblems}
\end{figure}

\begin{figure}[tb]
\includegraphics[width=\textwidth]{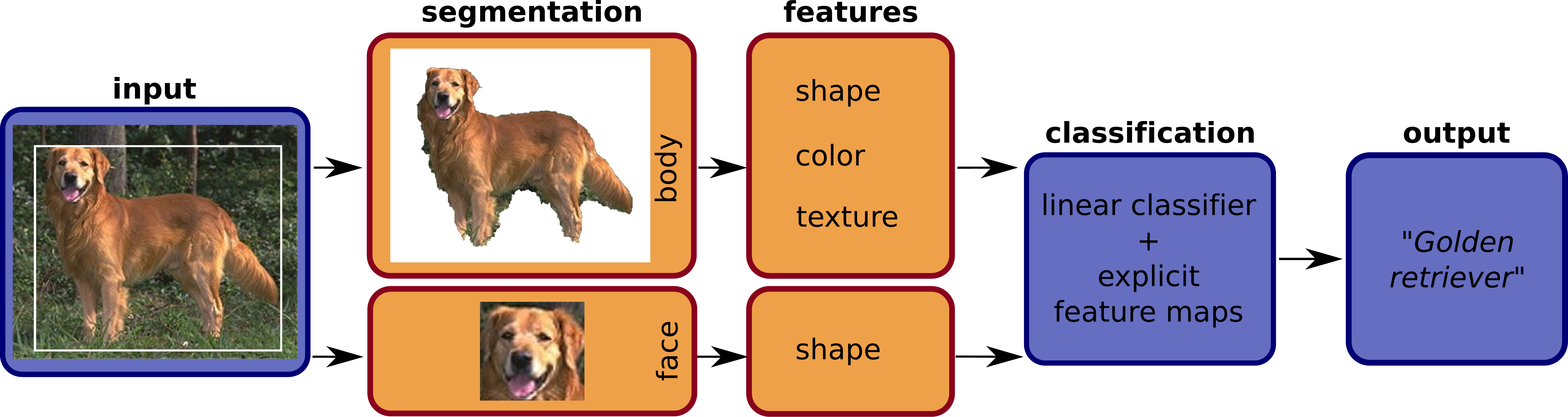}
\caption{Main algorithmic steps of our approach.}
\label{fig:overview}
\end{figure}

Within the last years, impressive success was achieved for classifying categories in real-world scenarios~\cite{Lazebnik06:BBF,Vedaldi09:MKO,Kapoor10:GPO,pascal-voc-2012}.
However, it is still an open problem how to reliably differentiate between visually similar classes, a task which is also known as \emph{fine-grained object classification}~\cite{yao12-caa,zhang12-ppk, chai12-ttc, parkhi12-cd}. 
In the following, we tackle fine-grained classification in the area of dog categorization. 
As can be seen in Figure \ref{fig:finegrainedproblems} for some example images, even for a human, this task is often hard to solve without expert knowledge. 
The goal of the challenge is to perform several visual recognition tasks with training and test data provided by the ImageNet~\cite{denk09-ilh} database. The challenge offers three different tasks: Classification, localization and fine-grained object classification.
Our team tackled task 3 of the challenge (ILSVRC 2012) related to fine-grained object classification and differentiating between different dog breeds.
We built a final classification system relying on three key ingredients: 
\begin{enumerate}
 \item a simple yet efficient part detector together with background elimination using a graph-based segmentation approach, and
 \item the combination of different feature types to capture different aspects of objects, namely shape, color, and texture,
 \item a linear classifier with an efficient kernel approximation to ensure computation times within a few hours even for this large-scale dataset.
\end{enumerate}
Details for every step follow in subsequent sections and an overview is given in \figurename~\ref{fig:overview}.

\section{Part-based and segmentation-based feature extraction}

In the following, we briefly describe the main algorithmic steps of our approach, which are mainly based on state-of-the-art segmentation and
feature extraction techniques.

\begin{figure}[b]
        \centering
        \begin{subfigure}[b]{0.35\textwidth}
                \centering
\includegraphics[width=0.9\textwidth]{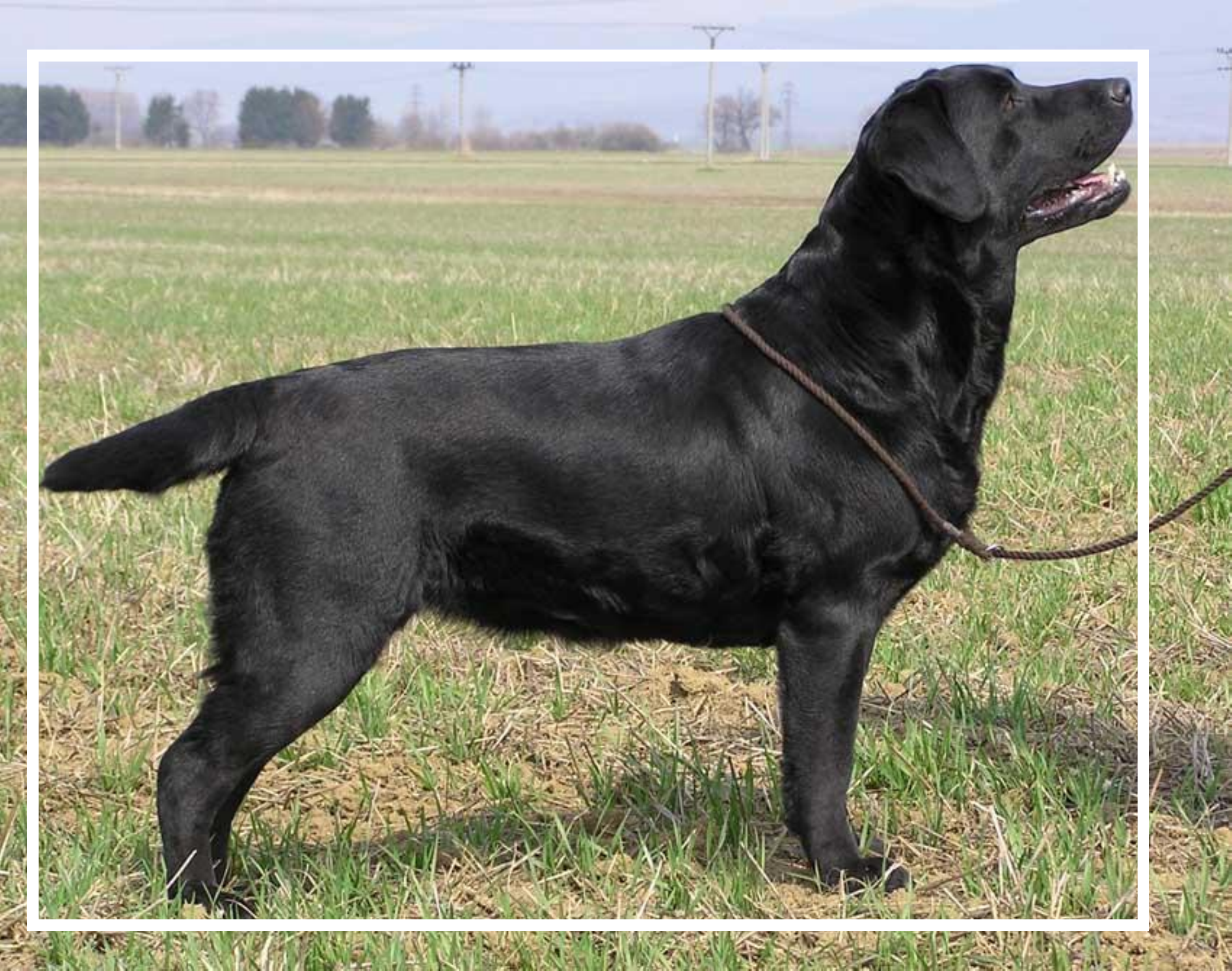}
                \caption{Input image with bounding box}
        \end{subfigure}
        \begin{subfigure}[b]{0.35\textwidth}
                \centering
\includegraphics[width=0.9\textwidth]{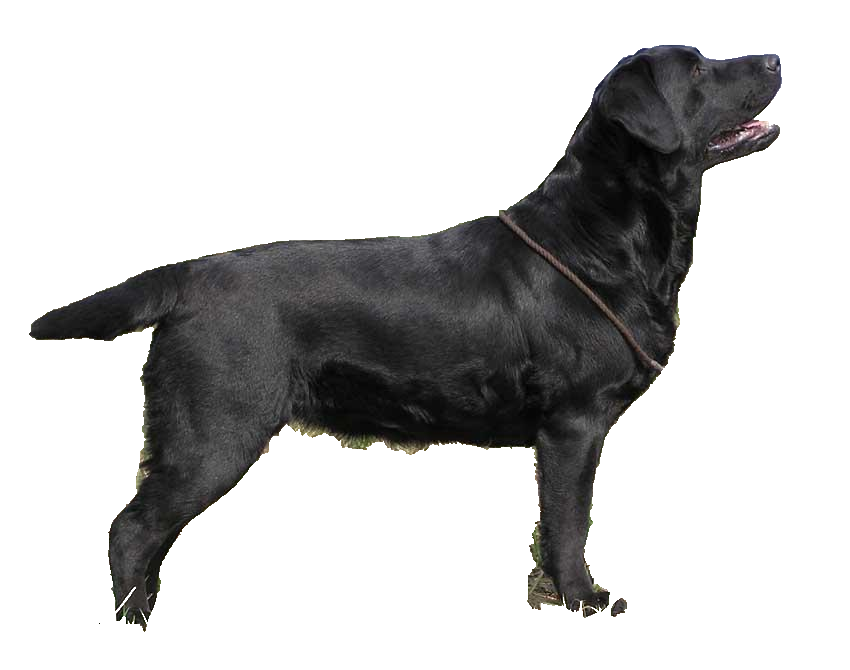}
                \caption{Result after applying Grabcut}
        \end{subfigure}
\caption{Background removal using Grabcut and the provided bounding box}
\label{fig:grabcut}
\end{figure}

\paragraph{Segmentation and foreground extraction}
Background clutter present in the images might interfere classification. We therefore
apply Grabcut~\cite{rother04-gif} to all images to consider relevant foreground regions only.
For Grabcut, a background color model was trained on the pixels outside of 
the provided bounding box, whereas a foreground color model was trained on 
pixels inside the bounding box. This initial bounding box segmentation is 
then refined using iterated graph cuts (\figurename~\ref{fig:grabcut}).

\begin{figure}[tb]
        \centering
        \begin{subfigure}[b]{0.3\textwidth}
                \centering
\includegraphics[width=\textwidth]{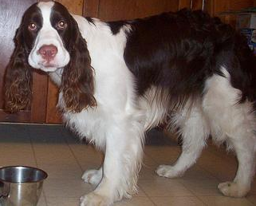}
                \caption{input image}
        \end{subfigure}
        \begin{subfigure}[b]{0.3\textwidth}
                \centering
\includegraphics[width=\textwidth]{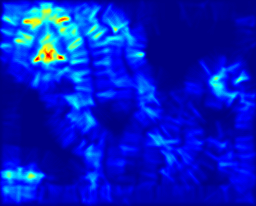}
                \caption{Hough circle transformation}
        \end{subfigure}
        \begin{subfigure}[b]{0.3\textwidth}
                \centering
\includegraphics[width=\textwidth]{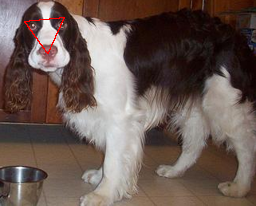}
                \caption{detected face}
        \end{subfigure}
\caption{Example of our dog face detection. First, the Hough circle transformation is computed and we then we search for a triangle with high circle activations in the image.}
	\label{fig:dogfacedetection}
\end{figure}

\begin{figure}[tb]
        \begin{subfigure}[b]{\textwidth}
                \centering
\includegraphics[height=0.13\textheight]{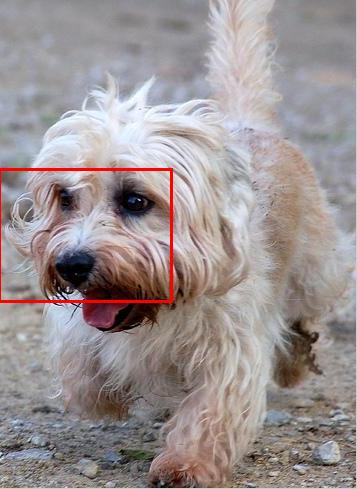}
\includegraphics[height=0.13\textheight]{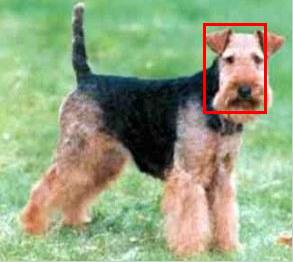}
\includegraphics[height=0.13\textheight]{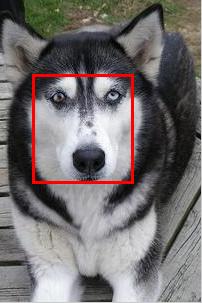}
\includegraphics[height=0.13\textheight]{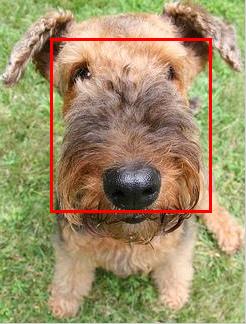}
\includegraphics[height=0.13\textheight]{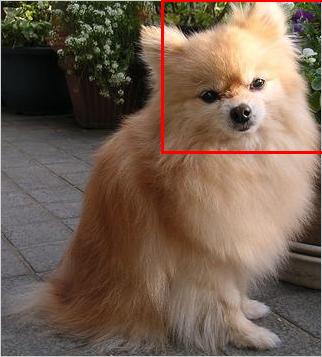}
                \caption{successfully detected faces}
		\end{subfigure}

        \begin{subfigure}[b]{\textwidth}
                \centering
\includegraphics[height=0.128\textheight]{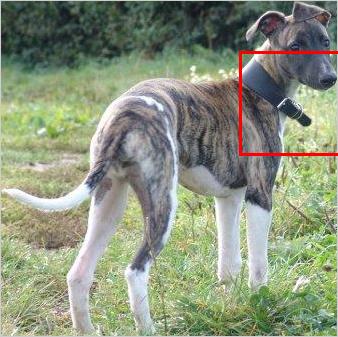}
\includegraphics[height=0.128\textheight]{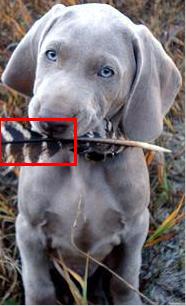}
\includegraphics[height=0.128\textheight]{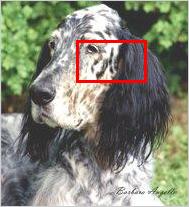}
\includegraphics[height=0.128\textheight]{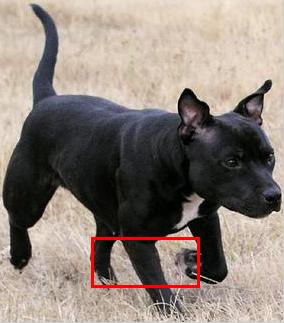}
\includegraphics[height=0.128\textheight]{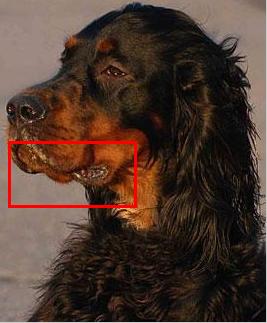}
                \caption{miss-detections}
		\end{subfigure}

\caption{Randomly selected results of our face detection algorithm. Even though we used a very simple algorithm a relatively large fraction of the images shows successful detection. But the algorithm fails if there is low contrast between the eyes and the face or if there is no frontal view of the face.}
	\label{fig:dogfacedetectionresults}
\end{figure}

\paragraph{Including part-based information}
Following state-of-the-art approaches~\cite{liu12-bcu,felzenszwalb10-odd}, we additionally extract part based information.
Since the global part constellations of dogs we observe in images have a large variation, the parts being most reliably detectable are the heads. 
Unfortunately, there is no annotation for these parts available in the data and we can not train a standard detector as done by~\cite{liu12-bcu}.
Therefore, we develop a simple head detector by 
applying a Hough circle transform to find both eyes and the nose, where we in particular search for three circles that compose a triangle.
Figure~\ref{fig:dogfacedetection} shows an example of the process.
With this approach, we are able to find dog heads in the images with a surprisingly high detection rate. 
Our detection approach does not work with dark fur, bad illumination conditions, and when the head is not in the picture.
Figure~\ref{fig:dogfacedetectionresults} shows some examples of successful and unsuccessful detections.
The detection results are used to extract an additional SIFT bag-of-words descriptor from the head region.

\paragraph{Feature extraction -- a kitchen sink approach}
For differentiating between hundreds of dog categories, many details matter.
Therefore, we represent images with a combination of different sources of information. 
The shape of objects is captured using a bag-of-words histogram of
opponent SIFT~\cite{sande10-ecd} descriptors that are densely sampled from the image.
In addition, we extract color information using color name~\cite{weijer09-lcn} histograms.
Finally, we compute local binary patterns~\cite{ojala02-mgr} to capture texture information, which
might be helpful for differentiating between different fur structures.
We add spatial information to every type of feature by extracting not a single feature per image
but a representation based on pyramid histograms~\cite{bosch07-rsw}.
All of these methods are standard techniques in the area of object recognition and showed to provide a good performance for various datasets and tasks.

\paragraph{Classification -- speed matters}
Images are represented by a combination of all previously described features, \textit{i.e.} simple concatenation of features instead
of a multiple kernel learning approach like the one used in~\cite{rodner2010multiple}.
For classification, we use a linear SVM learned with the \texttt{liblinear}~\cite{fan08-lll} package 
in a one-vs-all manner.
Due to the linearity of the classifier, learning and classification is extremely fast,
which makes it feasible for this large-scale dataset. However, the gain of speed has the
drawback of a diminished discriminative power. We overcome this drawback by utilizing homogeneous 
kernel maps~\cite{vedaldi12-eak} to approximate a $\chi^2$-kernel. With this combination, we are able to combine 
the speed of a linear SVM with the discriminative power of kernel-based methods.
In our case \texttt{liblinear} is able to train a model using the 20500 training examples
in less than 4 hours using 70GB RAM.

\section{Implementation details and parameters}
The dataset used in the challenge consists of 120 different breeds of dogs with different poses, background and illuminations.
It is divided into a training, validation and test set.
The training set contains of 20580 images, the validation set contains of 6000 images and the test set contains 12000 images of dogs.
Some example images can be seen in Figure~\ref{fig:finegrainedproblems}.
Bounding boxes are provided for all images, but ground truth labels are only provided for the training and validation set.
The measure used for evaluation is the mean average precision (mAP) as used in Pascal VOC 2012~\cite{pascal-voc-2012}.

Our system was programmed in Matlab with the help of several external libraries.
We used available implementations for 
color names\footnote{\url{http://lear.inrialpes.fr/people/vandeweijer/color_names.html}},
local binary patterns\footnote{\url{http://www.cse.oulu.fi/CMV/Downloads/LBPMatlab}},
grab cut\footnote{\url{http://opencv.org/}},
\texttt{liblinear}\footnote{\url{http://www.csie.ntu.edu.tw/~cjlin/liblinear/}},
SIFT as well as homogeneous kernel maps\footnote{\url{http://www.vlfeat.org/}}.

SIFT features were densely sampled. We used a 1000 word vocabulary for the whole object and a 500 word dictionary for the detected face region.
The local binary pattern features were extracted at a scale of 1, 2, and 4.
Furthermore, we applied a 3 level spatial pyramid for all features except the color name histograms, for which we used a 5 level spatial pyramid. 
For classification, a homogeneous kernel map of order 1 was used with $\gamma$ set to $0.5$. The parameter $C$ was set to $10$ for SVM.

%parameters:
%colorname - pyramid:5
%%sift - 1000 words
%lbp - pyramid 3, radius 1, 2 und 4
%face - sift, 500 words, 

\section{Experimental evaluation}

\begin{table}[tbp]
\centering
\begin{tabular}{lcc}
\toprule
feature & mAP & recognition rate\\
\midrule
SIFT (whole image) & 0.1933 & 0.2157\\
color names & 0.0475 & 0.0617\\
local binary patterns s=1 & 0.0529 & 0.0650\\
local binary patterns s=2 & 0.0699 & 0.0962\\
local binary patterns s=4 & 0.0729 & 0.1027\\
SIFT (head region) & 0.0729 & 0.0905\\
\midrule
combined & 0.2524 & 0.2757\\
\bottomrule
\end{tabular}
\caption{This table shows the results of single and combined features on the validation set of task 3.}
\label{tbl:results}
\end{table}

\begin{table}[position specifier]
\centering
\begin{tabular}{lc}
\toprule
feature & mAP \\
\midrule
SIFT + color names & 0.2090\\
SIFT + local binary patterns s=1 & 0.2141\\
SIFT + local binary patterns s=2 & 0.2193\\
SIFT + local binary patterns s=4 & 0.2153\\
SIFT + SIFT-faces & 0.2082\\
\midrule
all combined & 0.2524\\
\bottomrule
\end{tabular}
\caption{Results of pairs of features on the validation set of task 3. One can see that every feature contributes to the final result.}
\label{tbl:resultspairs}
\end{table}

\begin{figure}[tb]
        \begin{subfigure}[b]{0.19\textwidth}
                \centering
\includegraphics[height=0.7\textwidth]{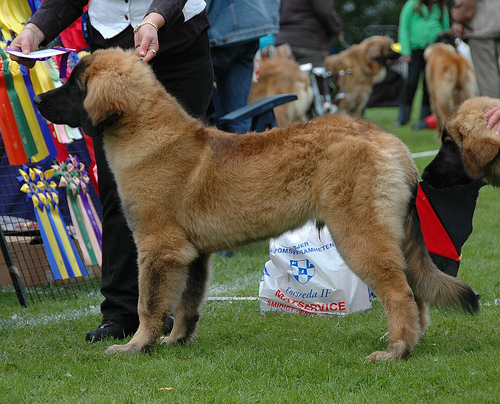}
                \caption{Leonberg\\AP: 0.7974}
        \end{subfigure}
        \begin{subfigure}[b]{0.19\textwidth}
                \centering
\includegraphics[height=0.7\textwidth]{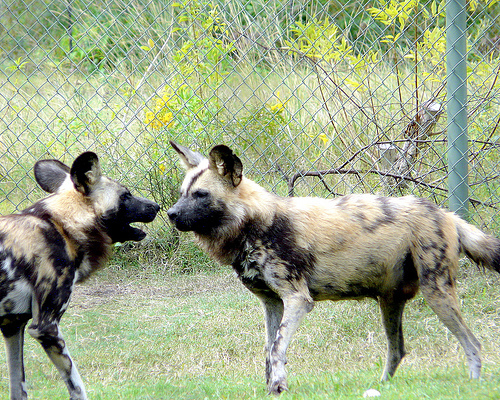}
                \caption{African hunting\\AP: 0.7344}
        \end{subfigure}
        \begin{subfigure}[b]{0.19\textwidth}
                \centering
\includegraphics[height=0.7\textwidth]{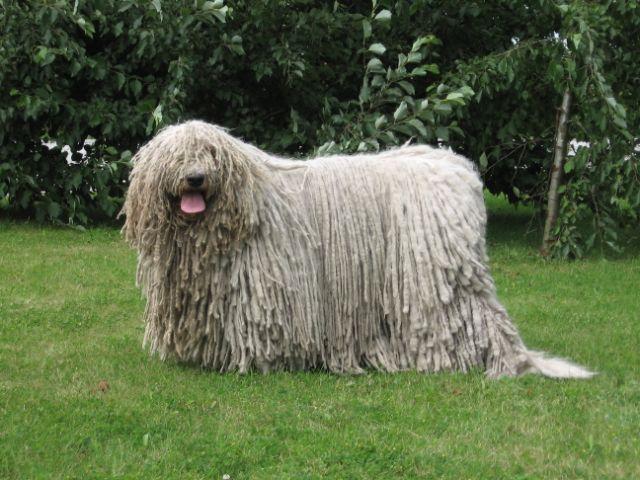}
                \caption{Komondor\\AP: 0.6158}
        \end{subfigure}
        \begin{subfigure}[b]{0.19\textwidth}
                \centering
\includegraphics[height=0.7\textwidth]{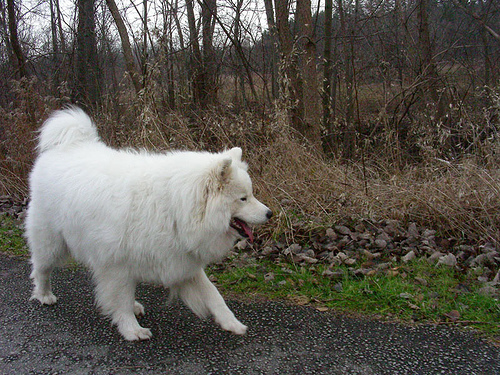}
                \caption{Samoyed\\AP: 0.6072}
        \end{subfigure}
        \begin{subfigure}[b]{0.19\textwidth}
                \centering
\includegraphics[height=0.7\textwidth]{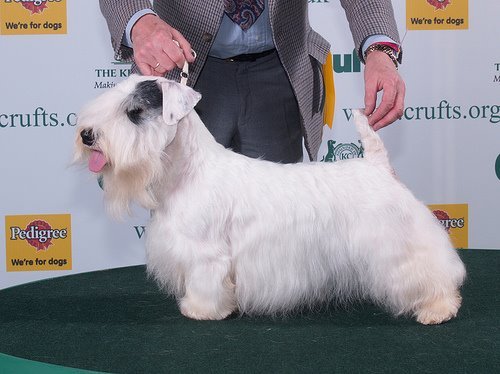}
                \caption{Sealyham terrier\\AP: 0.5765}
        \end{subfigure}
\caption{The 5 classes with the highest AP on the validation dataset.}
\label{fig:top5}
\end{figure}

\begin{figure}[tb]
        \begin{subfigure}[b]{0.19\textwidth}
                \centering
\includegraphics[height=0.7\textwidth]{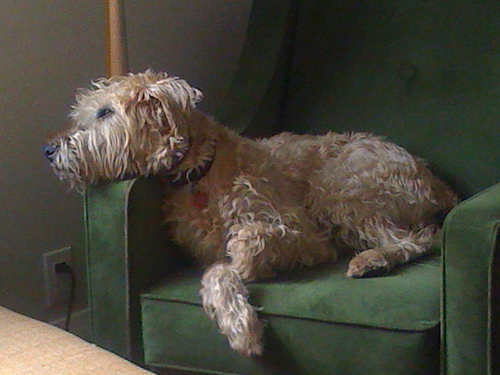}
                \caption{So. wh. Terrier\\AP: 0.0349}
        \end{subfigure}
        \begin{subfigure}[b]{0.19\textwidth}
                \centering
\includegraphics[height=0.7\textwidth]{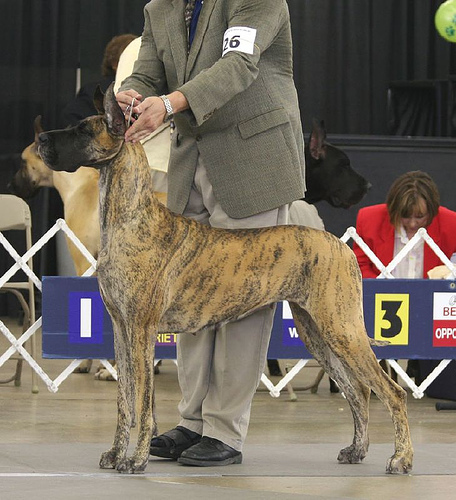}
                \caption{Great Dane\\AP: 0.0562}
        \end{subfigure}
        \begin{subfigure}[b]{0.19\textwidth}
                \centering
\includegraphics[height=0.7\textwidth]{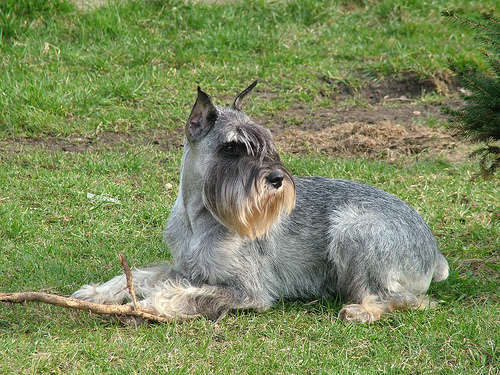}
                \caption{Std. Schnauzer\\AP: 0.0652}
        \end{subfigure}
        \begin{subfigure}[b]{0.19\textwidth}
                \centering
\includegraphics[height=0.7\textwidth]{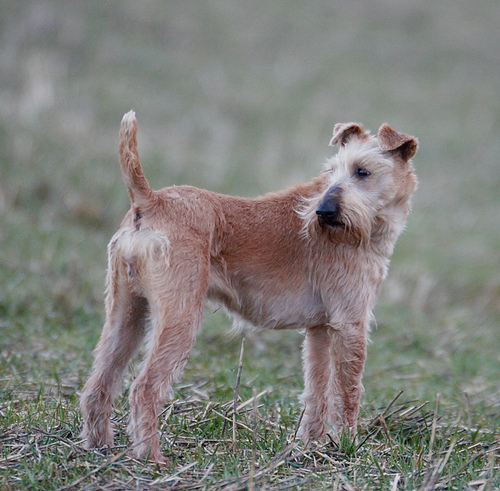}
                \caption{Irish terrier\\AP: 0.0776}
        \end{subfigure}
        \begin{subfigure}[b]{0.19\textwidth}
                \centering
\includegraphics[height=0.7\textwidth]{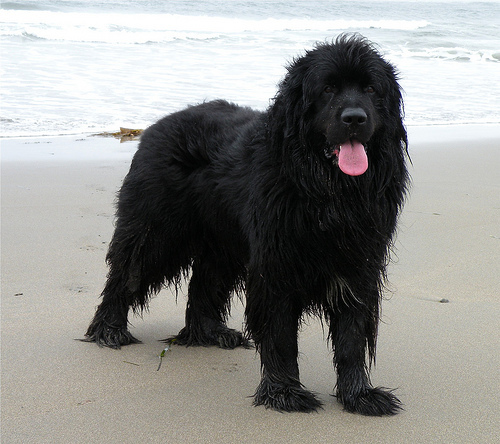}
                \caption{Newfoundland\\AP: 0.0782}
        \end{subfigure}
\caption{The 5 classes with the lowest AP on the validation dataset.}
\label{fig:bottom5}
\end{figure}

\paragraph{Results on the validation set}
Table \ref{tbl:results} shows the results of the individual features and the result of combining all the features.
The mean average precision achieved was 0.2524 on the validation set and 0.2459 on the test set.
Table \ref{tbl:resultspairs} shows results using pairs of features. Here we can see that each of the extracted features contributes to the final result.
Figure \ref{fig:top5} and \ref{fig:bottom5} show the 5 classes with the highest and lowest mAP, respectively. 

\begin{figure}[tb]

        \begin{subfigure}[b]{0.19\textwidth}
                \centering
\includegraphics[height=0.7\textwidth]{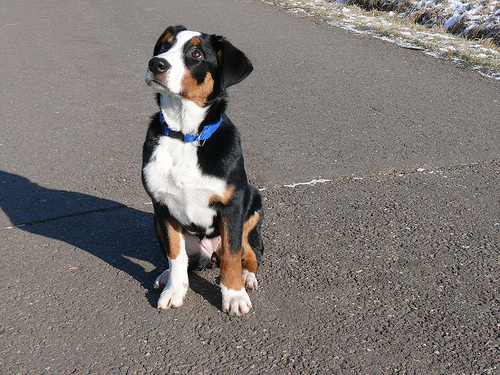}
\includegraphics[height=0.7\textwidth]{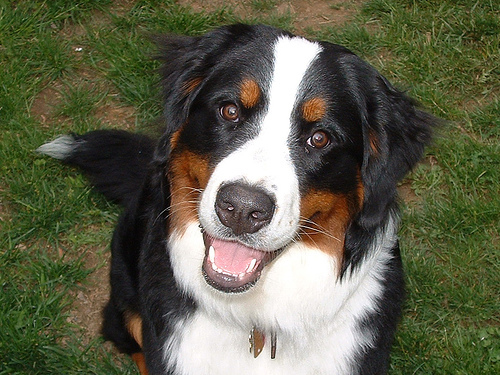}
                \caption{Appenzeller\\ as\\ Bernese Mount.}
        \end{subfigure}
        \begin{subfigure}[b]{0.19\textwidth}
                \centering
\includegraphics[height=0.7\textwidth]{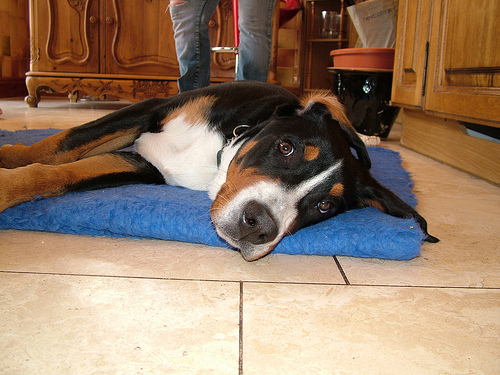}
\includegraphics[height=0.7\textwidth]{img/confused/ILSVRC2012_val_00040427.JPEG}
                \caption{G. Swiss Mount.\\as\\ Bernese Mount.}
        \end{subfigure}
        \begin{subfigure}[b]{0.19\textwidth}
                \centering
\includegraphics[height=0.7\textwidth]{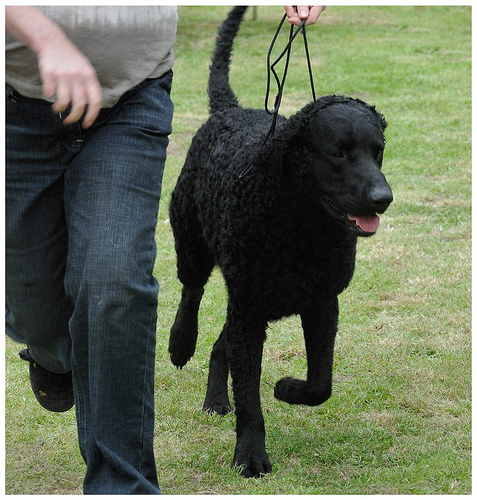}
\includegraphics[height=0.7\textwidth]{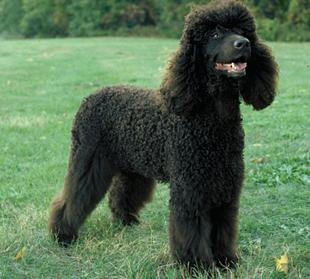}
                \caption{Curl. Retriever\\ as\\ Ir. W. Spaniel}
        \end{subfigure}
        \begin{subfigure}[b]{0.19\textwidth}
                \centering
\includegraphics[height=0.7\textwidth]{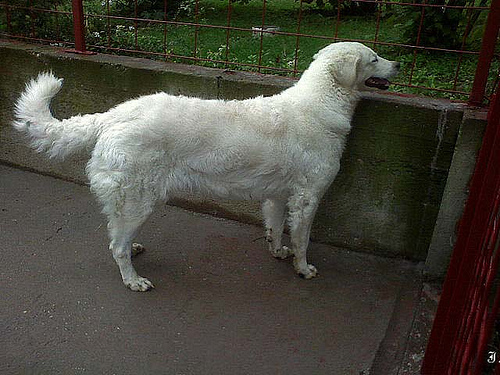}
\includegraphics[height=0.7\textwidth]{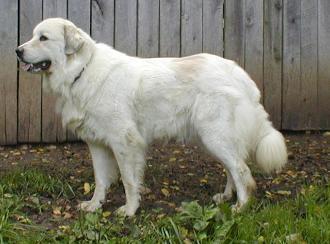}
                \caption{Kuvasz\\ as\\ Great Pyrenees}
        \end{subfigure}
        \begin{subfigure}[b]{0.19\textwidth}
                \centering
\includegraphics[height=0.7\textwidth]{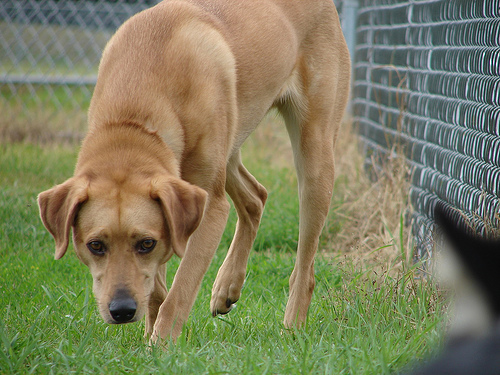}
\includegraphics[height=0.7\textwidth]{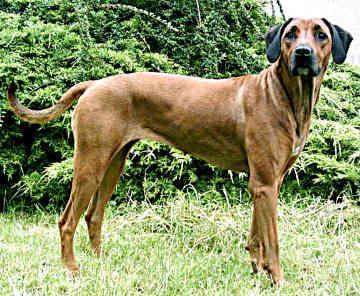}
                \caption{Vizsla, Hung.\\ as\\ Rhod. Ridgeback}
        \end{subfigure}
\caption{The 5 pairs of classes that have been confused the most often on the validation dataset.}
	\label{fig:confused}
\end{figure}

\begin{figure}[tb]
	\centering
\includegraphics[width=0.5\textwidth]{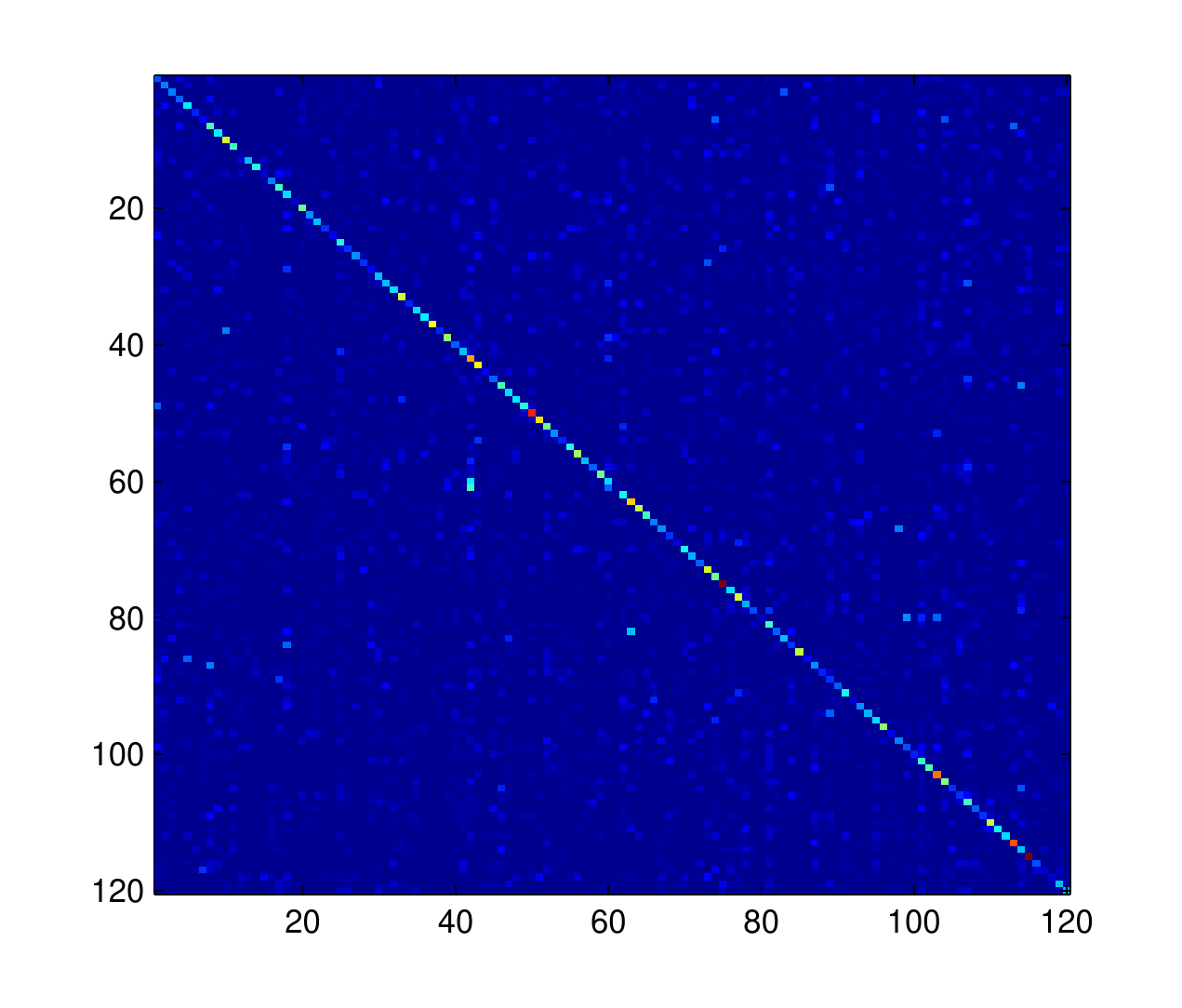}
	\caption{The confusion matrix computed using the validation dataset.}
	\label{fig:confusionmat}
\end{figure}

To create Figure~\ref{fig:confused}, we analyzed the highest entries of the confusion matrix that can be seen in Figure \ref{fig:confusionmat}.
These classes are indeed very similar and even a human annotator without the necessary expert knowledge would struggle to distinguish these breeds.

\paragraph{Results on the test set}
Our approach obtained a recognition rate of $0.245897$ in the 2012 challenge, whereas the methods of the XRCE/INRIA team and the ISI team were able to achieve a recognition of 0.309932 and 0.322524, respectively. 
However, participating in the challenge was an interesting experience and our method shows the importance of feature combination and most importantly discriminative object part detections. Therefore, our results might help to boost the performance of other approaches.

% worst
% id	12		6		92		69		29
% AP	0.0349	0.0562	0.0652	0.0776	0.0782
% rR	0.02	0.12	0.06	0.08	0.1

% most confused
%	61	60	82	80	46
%	42	42	63	99	114
%	18	15	13	11	10

% most confused - symmetric
%	60	61	46	17	67
%	42	42	114	89	98
%	21	18	15	15	14

\section{Conclusion}
In this paper, we presented our approach to fine-grained image classification of our entry for the ImageNet Fine-grained Challenge 2012.
Our approach combines multiple cues, automatic background removal, and a simple part detector. 
We could show that each step leads to an improvement in recognition rate resulting in a good performance on the Stanford dog dataset.

Nevertheless, the dataset is very challenging and our algorithm is not yet able to distinguish between really similar looking classes.
Therefore, future work will be focused on how to extract fine-grained localized features that distinguish between these breeds.
Furthermore, more care has to be taken when handling varying poses. This could be done by either detection of multiple parts, calculating a feature representation with
zero intra-class variance~\cite{bodesheim2013kernel}, or by applying some kind of pose normalization to decrease the variation.
Global descriptors capturing more details of the underlying local features, like the fisher vectors used by the other competitors, would probably also lead to an improved result.

\bibliographystyle{plain}
\bibliography{papers}

\end{document}